\documentclass{article}

\usepackage{PRIMEarxiv}

\usepackage[utf8]{inputenc} 
\usepackage[T1]{fontenc}    
\usepackage{hyperref}       
\usepackage{url}            
\usepackage{booktabs}       
\usepackage{amsfonts}       
\usepackage{nicefrac}       
\usepackage{microtype}      
\usepackage{lipsum}
\usepackage{fancyhdr}       
\usepackage{graphicx}       
\graphicspath{{media/}}     

\usepackage{graphicx}
\usepackage{latexsym}
\usepackage{hyperref}

\usepackage{times}
\usepackage{booktabs}
\usepackage{float}
\usepackage{tabularx, multirow, multicol}
\usepackage{amsfonts,amssymb}

\title{Topic-Controllable Summarization:\\ Topic-Aware Evaluation and Transformer Methods}

\author{
  Tatiana Passali, Grigorios Tsoumakas\\
  School of Informatics \\
  Aristotle University of Thessaloniki \\
  Greece\\
  \texttt{\{scpassali, greg\}@csd.auth.gr} \\
}

\begin{document}
\maketitle

\begin{abstract}
Topic-controllable summarization is an emerging research area with a wide range of potential applications. However, existing approaches suffer from significant limitations. For example, the majority of existing methods built upon recurrent architectures, which can significantly limit their performance compared to more recent Transformer-based architectures, while they also require modifications to the model's architecture for controlling the topic. At the same time, there is currently no established evaluation metric designed specifically for topic-controllable summarization. This work proposes a new topic-oriented evaluation measure to automatically evaluate the generated summaries based on the topic affinity between the generated summary and the desired topic. The reliability of the proposed measure is demonstrated through appropriately designed human evaluation. In addition, we adapt topic embeddings to work with powerful Transformer architectures and propose a novel and efficient approach for guiding the summary generation through control tokens. Experimental results reveal that control tokens can achieve better performance compared to more complicated embedding-based approaches while also being significantly faster.
\end{abstract}


\section{Introduction}
Neural abstractive summarization models have matured enough to consistently produce high quality summaries~\cite{See2017GetNetworks,Dong2019UnifiedGeneration,zhang2020,lewis2020bart}. Building on top of this significant progress, an interesting challenge is to go beyond delivering a generic summary of a document, and instead produce a summary that focuses on specific aspects that pertain to the user's interests. For example, a financial journalist may need a summary of a news article to be focused on specific financial terms, like cryptocurrencies and inflation. 

This challenge has been recently addressed by \textit{topic-controllable summarization} techniques~\cite{Krishna2018,frermann2019inducing, bahrainian2022newts}. However, the automatic evaluation of such techniques remains an open problem. Existing methods use the typical ROUGE score~\cite{Lin2004} for measuring summarization accuracy and then employ user studies to qualitatively evaluate whether the topic of the generated summaries matches the users' needs~\cite{Krishna2018,bahrainian2021cats}. Yet, ROUGE can only be used for measuring the quality of the summarization output and cannot be readily used to capture the topical focus of the text. Some early steps have been made in this direction by~\cite{bahrainian2022newts}, employing a latent Dirichlet allocation (LDA) model to evaluate the topical focus of a summary. However, this serves as a simple indicator of the presence of the topic and cannot be directly used as an evaluation metric as it cannot be easily interpreted across different documents. Therefore, there is no clear way to automatically evaluate such approaches, since there is no evaluation measure designed specifically for topic-controllable summarization. 

At the same time, the majority of existing models for topic-controllable summarization either incorporate topic embeddings into the model's architecture~\cite{Krishna2018,frermann2019inducing} or modify the attention mechanism~\cite{bahrainian2021cats, Lu2024}. Since these approaches are restricted to very specific neural architectures, it is not straightforward to use them with any summarization model. Even though control tokens have been shown to be effective and efficient for controlling the output of a model for entity-based summarization~\cite{fan2018controllable,he2020ctrlsum,chan2021controllable} without any modification to its codebase, there have been limited efforts for topic-controllable summarization~\cite{bahrainian2021cats}.

Motivated by these observations, we propose a topic-aware evaluation measure for quantitatively evaluating topic-controllable summarization methods in an objective way, without involving expensive and time-consuming user studies. In particular, we propose calculating a summary representation of different topics and then calculating the cosine similarity between the generated summaries and the prototype topic vectors in relation with all the possible topics. The proposed measure assumes the existence of a pre-defined set of topics and thus can be easily adapted to any set of different topics. The effectiveness and reliability of the proposed measure are demonstrated through appropriately designed human evaluation. 

In addition, we extend prior work on topic-controllable summarization by adapting topic embeddings~\cite{Krishna2018} from the traditional RNN architectures to the more recent and powerful Transformer architecture~\cite{Vaswani2017}. However, as shown in experimental evaluation, this approach suffers from significant limitations e.g., slow inference. To this end, we propose a novel control token approach via three different approaches: i) \textit{prepending} the thematic category, ii) \textit{tagging} the most representative tokens of the desired topic, and iii) using both prepending and tagging. For the tagging-based method, given a topic-labeled collection, we extract keywords that are semantically related to the topic that the user requested and then employ special tokens to tag them before feeding the document to the model. We also demonstrate that control tokens can successfully be applied to zero-shot topic-controllable summarization, while at the same time being significantly faster than the embedding-based formulation and can be effortlessly combined with any neural architecture.

Our contributions can be summarized as follows:
\begin{itemize}
    \item We propose a topic-aware measure to quantitatively evaluate topic-oriented summaries, validated by a user study.
    \item We develop topic-controllable Transformers summarization methods.
    \item We provide an extensive empirical evaluation of the proposed methods, including an investigation of a zero-shot setting.
\end{itemize}

The rest of this paper is organized as follows. In Section~\ref{sec:related_work} we review existing related literature on controllable summarization. In Section~\ref{sec:metric} we introduce the proposed topic-aware measure, while in Section~\ref{sec:methods} we present the proposed methods for topic-controllable summarization. In section~\ref{sec:experiments} we discuss the experimental results. Finally, conclusions are drawn and interesting future research directions are given in Section~\ref{sec:conclusion}.

\section{Related Work}
\label{sec:related_work}
In this section, we first present an overview of related work in controllable text-to-text generation. Then, we discuss methods and models for improving abstractive summarization by incorporating topical information. Finally, we review techniques for topic control in neural abstractive summarization.

\subsection{Controllable text-to-text generation} Controllable summarization belongs to the broader field of controllable text-to-text generation~\cite{liu2021dexperts,pascual2021plug}. Several approaches for controlling the model's output exist, either using embedding-based approaches~\cite{Krishna2018,frermann2019inducing}, prepending information using special tokens~\cite{fan2018controllable,he2020ctrlsum} or using decoder-only architectures~\cite{liu2021dexperts}. Controllable summarization methods can influence several aspects of a summary, including its topic~\cite{Krishna2018,frermann2019inducing,bahrainian2021cats}, length~\cite{liu2020,chan2021controllable}, style~\cite{fan2018controllable}, and the inclusion of named entities~\cite{fan2018controllable,he2020ctrlsum,chan2021controllable}. Despite the importance of controllable summarization, there are still limited datasets for this task, including ASPECTNEWS~\cite{ahuja2022aspectnews} for extractive aspect-based summarization, EntSUM~\cite{maddela2022entsum} for entity-based summarization and NEWST~\cite{bahrainian2022newts} for topic-aware summarization.


\subsection{Improving summarization using topical information}

The integration of topic modeling into summarization models has been initially used in the literature to improve the quality of existing state-of-the-art models.
\cite{Ailem2019Topic} enhance the decoder of a pointer-generator network using the information of the latent topics that are derived from LDA. 
Similar methods have been applied by~\cite{wang2020friendly} using Poisson Factor Analysis (PFA) with a plug-and-play architecture that uses topic embeddings as an additional decoder input based on the most important topics from the input document. \cite{liu2021enhancing} propose to enhance summarization models using an Extreme Multi-Label Text Classification model to improve the consistency between the underlying topics of the input document and the summary, leading to summaries of higher quality.  \cite{zhu2021twag} use a topic-guided abstractive summarization model for Wikipedia articles leveraging the topical information of Wikipedia categories. Even though~\cite{wang2020friendly} refer to the potential of controlling the output conditioned on a specific topic, all the aforementioned approaches are focused on improving the accuracy of existing summarization models instead of influencing the summary generation towards a particular topic.

\subsection{Topic-control in neural abstractive summarization} Some steps towards controlling the output of a summarization model conditioned on a thematic category have been made by~\cite{Krishna2018,frermann2019inducing}, who proposed embedding-based controllable summarization models on top of the pointer generator network~\cite{See2017GetNetworks}. \cite{Krishna2018} integrate the topical information into the model as a topic vector, which is then concatenated with each of the word embeddings of the input text. \cite{bahrainian2021cats} propose to incorporate the topical information into the attention mechanism of the pointer generator network, using an LDA model.

With the advancements in Transformer architecture, Large Language Models (LLMs) such as GPT-3~\cite{brown2020language} and LLaMA~\cite{touvron2023llama} can also be employed for this task. For example, \cite{bahrainian2022newts} employ different prompting techniques to control the topic of the summary. Some steps towards integrating the Transformer architecture into topic-controllable summarization have been made by~\cite{Lu2024} who employ contextual embeddings within a constrained attention mechanism. In addition,~\cite{zesheng2023topic} adopt a topic-aware graph network to generate representations of topic nodes that are then ingested into the decoder of the summarization model to generate a topic-oriented summary.

However, a significant limitation of the majority of these approaches is that they require modifications to the architecture of the summarization model. Our work adapts the embedding-based paradigm to Transformers~\cite{Vaswani2017}, and employs control tokens, which can be applied effortlessly and efficiently to any model architecture as well as to the zero-shot setup.

\section{Topic-aware Evaluation Measure}
\label{sec:metric}
We propose a new topic-aware measure, called Summarization Topic Affinity Score (STAS), to evaluate the generated summaries according to their semantic similarity with the desired topic. STAS assumes the existence of a predefined set of topics $\mathcal{T}$, and that each topic, $t \in \mathcal{T}$, is defined via a set, $\mathcal{D}_t$, of relevant documents.

STAS is computed on top of vector representations of topics and summaries. Note that several options exist for extracting such representations, ranging from simple bag-of-words models to sophisticated language models like BERT~\cite{devlin2019bert}.  However, for simplicity, this work uses tf-idf vector representations to demonstrate that even with a simple representation, STAS can be successfully used for evaluating topic-controllable summarization models. More specifically, we employ the tf-idf model, where idf is computed across all documents $\bigcup_{t \in \mathcal{T}} \mathcal{D}_t$. Note that the use of idf allows us to weigh down common words that typically do not contain any important information about the topic of interest.

\begin{figure*}[t]
    \centering
    \includegraphics[width=0.5\textwidth]{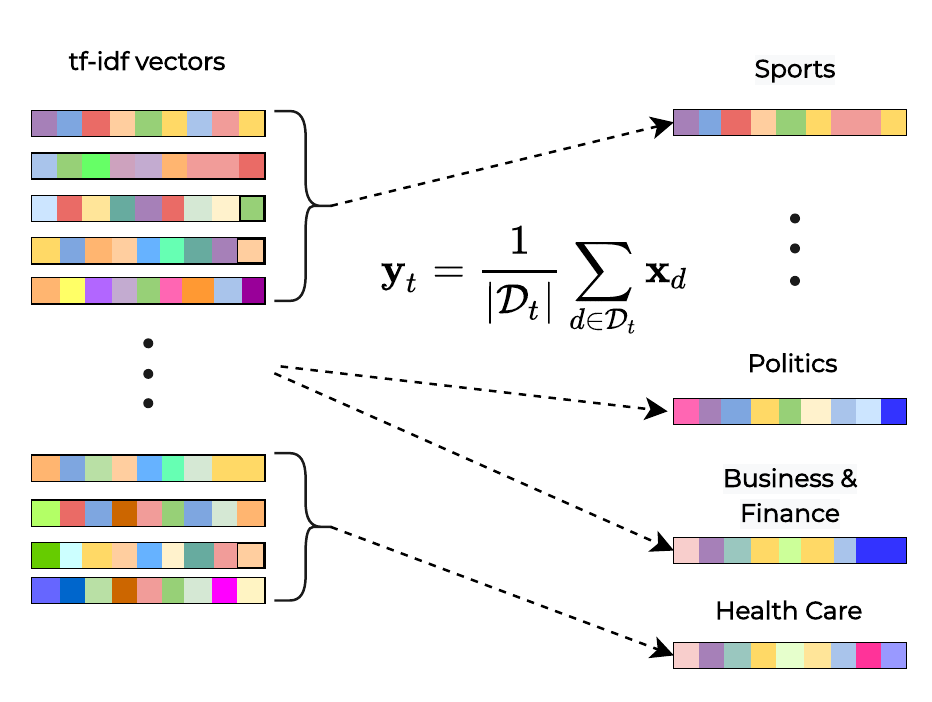}
    \caption{Obtaining representative words, given a topic-assigned document collection. First, we calculate vector representations for each document. Then, documents of the same topic are grouped and their vector representations is averaged. Finally, we obtain the words with top $N$ scores.
    }
    \label{fig:topical_vector_extraction}
\end{figure*}

For each topic $t$, we compute a topic representation $\mathbf{y}_t$ by averaging the vector representations, $\mathbf{x}_d$, of each document $d \in \mathcal{D}_t$ (see Fig.~\ref{fig:topical_vector_extraction}): 
\begin{equation}
    \mathbf{y}_t = \frac{1}{|\mathcal{D}_t|} \sum_{d \in \mathcal{D}_t} \mathbf{x}_d.
\end{equation}
The representation of the predicted summary, $\mathbf{y}_s$, is computed using the same model too, so as to lie in the same vector space with the topic vectors. 

To calculate STAS, we compute the cosine similarity between the representations of the summary, $\mathbf{y}_s$, and the desired topic, $\mathbf{y}_t$, divided by the maximum cosine similarity between the representations of the summary and all topics:

\begin{equation}
STAS(\mathbf{y}_s, \mathbf{y}_t) = \frac{s(\mathbf{y}_s, \mathbf{y}_t)}{\max\limits_{z \in T} \{ s(\mathbf{y}_s, \mathbf{y}_{z}) \}},
\end{equation}
where $s(\mathbf{y}_s, \mathbf{y}_t)$ indicates the cosine similarity between the two vectors $\mathbf{y}_s$ and $\mathbf{y}_t$, computed as follows:

\begin{equation}
s(\mathbf{y}_t, \mathbf{y}_s) = \frac{\mathbf{y}_t \mathbf{y}_s}{\|\mathbf{y}_t\| \|\mathbf{y}_s\|}.
\end{equation} 

Summaries that are similar to the requested topic receive high STAS values, while those that are dissimilar receive low STAS values. It is worth noting that cosine similarity values can differ significantly across topics, due to the varying overlap between the common words that appear in the summaries and the different topics. Dividing by the maximum value takes this phenomenon into account, leading to comparable STAS values across different topics and allows for normalizing the similarity over the dominant topic.

In addition, STAS can be an effective measure even when more than one dominant topic is discussed in the document, since it avoids distributing the similarity over topics that might appear to a smaller degree in a document, while allowing for normalizing the similarity over the dominant topics. For example, if the document contains two dominant topics, we expect that STAS will be near 1 for both dominant topics, while topics that appear to a smaller degree in a document will not affect the measure.

\section{Topic Control with Transformers}
\label{sec:methods}
In this section, we present the proposed topic-controllable summarization methods that fall into two different categories: a) incorporating topic embeddings into the Transformer architecture and b) employing control tokens before feeding the input to the model. Note that similar to the STAS measure, for all the proposed methods, we assume the existence of a predefined set of topics where each topic is represented from a set of relevant documents.

\subsection{Topic Embeddings}
Following other embedding-based methods for topic-controllable summarization~\cite{Krishna2018, frermann2019inducing}, we adapt a topic-aware pointer generator to work with Transformer-based architectures. As described in Section~\ref{sec:related_work}, \cite{Krishna2018} generate topic-oriented summaries by concatenating topic embeddings, which are represented as one-hot encoding vectors, to the token embeddings of a pointer generator network~\cite{See2017GetNetworks}. 
The topic embeddings are represented as one-hot encoding vectors with a size equal to the total number of topics. During training, the model takes as input the corresponding topic embedding along with the input document.

However, this method cannot be directly applied to pre-trained Transformer-based models due to the different shapes of the initialized weights of the word and position embeddings. Unlike RNNs, Transformer-based models are typically trained for general tasks and then fine-tuned with less data for more specific tasks like summarization. Thus, the architecture of a pre-trained model is already defined. Concatenating the topic embeddings with the contextual word embeddings of a Transformer-based model would require retraining the whole summarization model from scratch with the appropriate dimension. However, this would be computationally demanding as it would require a large amount of data and time.

Instead of concatenation, we propose to sum the topic embeddings, following the rationale of positional encoding, where token embeddings are summed with positional encoding representations to create an input representation that contains the position information. Instead of one-hot encoding embeddings, we use trainable embeddings allowing the model to optimize them during training. The topic embeddings have the same dimensionality as the token embeddings.

During training, we sum the trainable topic embeddings with token and positional embeddings and we modify the input representation as follows:

\begin{equation}
z_{i} = WE(x_{i}) + PE(i) + TE,
\end{equation}
where WE, PE and TE are the word embeddings, positional encoding and topic embeddings respectively, for token $x_{i}$ in position $i$.
During inference, the model generates the summary based on the trained topic embeddings, according to the desired topic.

\subsection{Control Tokens}
\label{sec:control_tokens}
We propose three different approaches to control the generation of the output summaries using control tokens: a) \textit{prepending} the thematic category as a special token to the document, b) \textit{tagging} with special tokens the representative terms for each topic, and c) combination of both control tokens.

There exist several controllable approaches that prepend information to the input source to influence the different aspects of the text such as the style~\cite{fan2018controllable} or the presence of a particular entity~\cite{he2020ctrlsum,chan2021controllable}. Even though this technique can be readily combined with topic controllable summarization, this direction has not been explored yet. We adapt these approaches to work with topic information by simply placing the desired thematic category at the beginning of the document. For example, prepending \textit{``Sports''} to the beginning of a document represents that we want to generate a summary based on the topic \textit{"Sports"}. During training, we prepend to the input the topic of the target summary, according to the training dataset. During inference, we also prepend the document with a special token according to the user's requested topic.

Going one step further, we propose another method for controlling the output of a model based on tagging the most representative terms for each thematic category using a special control token. The proposed method assumes the existence of a set of the most representative terms for each topic. Then, given a document and a requested topic, we employ a special token to tag the most representative terms before feeding the document to the summarization model, aiming to guide the summary generation towards this topic. We extract $N$ representative terms for each topic by considering the terms with the top $N$ scores in the corresponding topical vector, as extracted in Section~\ref{sec:metric}. Before training with a neural abstractive model, we pre-process each input document by surrounding the representative terms of its summary's topic with the special token [TAG]. This way, during training the model learns to pay attention to tagged words, as they are aligned with the topic of the summary.To influence the summary generation towards a desired topic during inference, the terms of this topic that appear inside the input document are again tagged with the special token.

Given the set of representative words for each topic, a document, and the desired topic, the tagging mechanism works as follows. All the words of the input document are lemmatized to their roots. Then, we identify the common words between the existing lemmatized tokens and the representative words for the desired topic. Finally, we tag each token of the input document with a special token, i.e., [TAG], only if the lemmatized form of this token is contained in the set of the most representative words for the corresponding topic. Some examples of representative terms for each topic are shown in Table~\ref{tab:most_common_words}.

\begin{table}
\centering
\label{tab:most_common_words} 
\caption{Representative terms for topics from 2017 KDD Data Science+Journalism Workshop~\cite{VOX_DATASET}.}
\begin{tabularx}{\columnwidth}{lX}
\toprule
\textbf{Topic} & \textbf{Terms} \\ \midrule
Politics    & policy, president, state, political, vote, law, country, election \\  \midrule
Sports  &   game, sport, team, football, fifa, nfl, player, play, soccer, league \\ \midrule
Health Care & patient, uninsured, insurer, plan, coverage, care, insurance \\ \midrule
Education & student, college, school, education, test, score, loan, teacher \\ \midrule
Movies & film, season, episode, show, movie, character, series, story \\ \midrule
Space & earth, asteroid, mars, comet, nasa, space, mission, planet \\ \midrule
\end{tabularx}

\end{table}

For example, suppose that we pre-process the sentence below, as a part of an input document, from which we aim to guide the generation towards the topic ``\textit{Business \& Finance}''. Following the aforementioned procedure, we will enclose with the special token [TAG], the words ``\textit{businesses}'', ``\textit{billion}'' and ``\textit{tax}'' since they belong to the set of the most representative words for the desired topic, as follows. 

\begin{quote}
   ``By one estimate, American individuals and [TAG]~\textbf{businesses}~[TAG] together spend 6.1 [TAG]~\textbf{billion}~[TAG] hours complying with the [TAG]~\textbf{tax}~[TAG] code every year.''

\end{quote}

\subsection{Topical Training Dataset}
\label{ssec:training_dataset}
All the aforementioned methods assume the existence of a training dataset, where each summary is associated with a particular topic. However, currently there are no existing large-scale training datasets for abstractive summarization that contain summaries according to the different topical aspects of the text. Thus, we adopt the approach of~\cite{Krishna2018} to compile and release a topic-oriented dataset.

More specifically, \cite{Krishna2018} create a topic-oriented dataset which contains new super-articles by combining two different articles of the original dataset and keeping the summary of only one of them. First, they extract BoW vector representations for each topic from the Vox dataset~\cite{VOX_DATASET}. Then, they compute the dot-product between the BoW representation of the summary and all the BoW topic representations. The topic with the highest similarity is assigned to the corresponding article, while articles with more than one dominant topic are discarded. All the topic-assigned articles are used to compile a temporary intermediate dataset. 

To create the final topic-oriented dataset, two articles $a_1$ and $a_2$ with different topics are randomly selected from the intermediate dataset. A new article $a'$ is created by sequentially selecting sentences from both articles. The new article $a'$ is assigned with the summary of one of the two selected articles and the same process is repeated to create a new article $a''$ that is assigned with the remaining summary. Then, the initially selected articles $a_1$ and $a_2$ are removed from the intermediate dataset. This process is continued until there are no articles in the intermediate dataset or all the remaining articles belong to the same topic. 

The final topic-oriented dataset consists of super-articles that discuss two distinct topics, but are assigned each time to one of the corresponding summaries. Therefore, the model learns to distinguish the most important sentences for the corresponding topic during training. Even though this procedure requires some additional effort, it allows us to effectively train our models on a topic-controllable setup. This dataset is used to fine-tune all the aforementioned methods.

\section{Empirical Evaluation}
\label{sec:experiments}
In this section, we present and discuss the experimental evaluation results. First, we introduce the experimental setup used for the evaluation, including the dataset generation procedure, the evaluation metrics, and employed deep learning architectures. Then, we proceed by presenting and discussing the experimental results

\subsection{Experimental Setup}
We use the following two datasets to evaluate the proposed models: (a) CNN/DailyMail and b) Topic-Oriented CNN/DailyMail. CNN/DailyMail is an abstractive summarization dataset with articles from CNN and DailyMail accompanied with human generated bullet summaries~\cite{hermann2015teaching}. We use the non-anonymized version of the dataset similar to~\cite{See2017c}. Topic-Oriented CNN/DailyMail is a synthetic version of CNN/Dailymail which contains super-articles of two different topics accompanied with the summary for the one topic.

To compile the topic-oriented CNN/DailyMail dataset any dataset that contains topic annotations can be used. Following~\cite{Krishna2018}, we also use the Vox Dataset~\cite{VOX_DATASET} which consists of 23,024 news articles of 185 different topical categories. We discarded topics with relatively low frequency, i.e. lower than 20 articles, as well as articles assigned to general categories that do not discuss explicitly a topic, i.e. ``\textit{The Latest}'', ``\textit{Vox Articles}'', ``\textit{On Instagram}'' and ``\textit{On Snapchat}''. 
%
After pre-processing, we end up with 14,312 articles from 70 categories out of the 185 initial topical categories. 


The final synthetic topic-oriented CNN/DailyMail consists of 132,766, 5,248, and 6,242 articles for training, validation, and test, respectively while the original CNN/DailyMail consists of 287,113, 13,368 and 11,490 articles. 

The Vox dataset is also used to extract the topic vector representations for the STAS measure. We use the tf-idf vectorizer provided by the Scikit-learn library~\cite{scikitlearn} to extract a vector representation for each document in the corpus. Then, all the representations of the same topic are averaged to extract the final vector representation for each topic. 


For the tagging-based method, all the words of the input document are lemmatized to their roots using NLTK~\cite{bird2006nltk}. Then, we tag the words between the existing lemmatized tokens and the representative words for the desired topic, based on the top-$N$=100 most representative terms for each topic.

For all the conducted experiments we employ a BART-large architecture~\cite{lewis2020bart}, which is a transformer-based model with a bidirectional encoder and an auto-regressive decoder. BART-large consists of 12 layers for both encoder and decoder and 406M parameters. We use the established parameters for the BART-large architecture and the implementation provided by Hugging Face~\cite{wolf2020transformers}. All the models are fine-tuned for 100,000 steps with a learning rate of  3×10$^{-5}$ and batch size 4, with early stopping on the validation set. We use PyTorch version 1.10 and Hugging Face version 4.11.0. All the models were trained using GPUs available in Google Colab, and in specific the NVIDIA T4 Tensor 16 GB GPU. The code and the compiled dataset are publicly available\footnote{\href{https://github.com/tatianapassali/topic-controllable-summarization}{https://github.com/tatianapassali/topic-controllable-summarization}}.

All methods were evaluated using both the well-known ROUGE~\cite{Lin2004} score, to measure the quality of the generated summary, as well as the proposed STAS measure.

\subsection{Results}
The evaluation results on the compiled topic-oriented dataset are shown in Table~\ref{tab:experimental_results}. Our results include the following models:
\begin{enumerate}
    \item  \textbf{PG} \cite{See2017GetNetworks} which is a generic pointer generator network, which is based on the RNN's architecture
    \item \textbf{Topic-Oriented PG}~\cite{Krishna2018} which is the topic-oriented pointer generator network also based on the RNN's architecture.
    \item \textbf{BART}~\cite{lewis2020bart} which is the generic BART model which is based on the Transformer-based architecture.
    \item \textbf{BART\textsubscript{emb}} which is the proposed topic-oriented embedding-based extension of BART.
    \item \textbf{BART\textsubscript{tag}} which is the proposed topic-oriented tagging-based extension of BART.
    \item \textbf{BART\textsubscript{pre}} which is the proposed topic-oriented prepending-based extension of BART.
    \item \textbf{BART\textsubscript{pre+tag}} which is the combination of the tagging and prepending extensions of BART.
\end{enumerate}


The experimental results reported in Table~\ref{tab:experimental_results} show that topic control methods perform significantly better compared to the corresponding baseline methods that do not take into account the topic requested by the user. Furthermore, the proposed BART-based formulation significantly outperforms the topic-oriented PG approach, regardless of the applied method (BART\textsubscript{emb}, BART\textsubscript{tag} or BART\textsubscript{pre}). The best results are obtained when the tagging and prepending methods are combined. The effectiveness of using topic-oriented approaches is further highlighted using STAS, since the improvements acquired when applying the proposed methods are much higher compared to the improvements in ROUGE score. The embedding and tagging methods lead to similar results 
(around 68.5\%), while the prepending method achieves better results (71.9\%). Finally, when we combine the tagging and prepending methods, we observe additional gains, outperforming all the evaluated methods with a 72.36\% STAS score. 

\begin{table}
\centering
\label{tab:experimental_results} 
\caption{Experimental results on the compiled topic-oriented dataset based on CNN/DailyMail dataset. We report $F_1$ scores for ROUGE-1 (R-1), ROUGE-2 (R-2) and ROUGE-L (R-L) and inference time for 100 articles. Time is reported in seconds.}
\begin{tabular}{ccccc|ccc}
\toprule
     & \textbf{R-1} & \textbf{R-2}   & \textbf{R-L} & \textbf{STAS (\%)} & \textbf{Control Tokens} &\textbf{ Inference} & \textbf{Total Time} \\ \midrule
PG & 26.8 & 9.2  & 24.5 & - & - & - & - \\ \midrule
BART  & 30.46 & 11.92 &  20.57 & 51.86  & - & 30.8 & 30.8 \\ \midrule
Topic-Oriented PG & 34.1 & 13.6  & 31.2 & - & - & - & -\\ \midrule
BART\textsubscript{tag} (Ours) & 39.30 & 18.06 & 36.67 & 68.42 & 7.1 & 32.0 & 39.1\\ \midrule 
BART\textsubscript{emb} (Ours) & 40.15 & 18.53 & 37.41 & 68.50  & - & 303.0 & 303.0 \\ \midrule
BART\textsubscript{pre} (Ours) & 41.58 & 19.55 & 38.74 & 71.90 & $<$0.1 & 30.9 & 30.9 \\ \midrule 
BART\textsubscript{pre+tag} (Ours) & \textbf{41.66} & \textbf{19.57} & \textbf{38.83} & \textbf{72.36} & 7.1 & 31.7 & 39.7 \\  \bottomrule
\end{tabular}

\end{table}

In addition, the inference time of all the methods that use control tokens is significantly smaller, improving the performance of the model by almost one order of magnitude. Indeed, all the control tokens approaches can perform inference on 100 articles in less than 40 seconds, while the embedding-based formulation requires more than 300 seconds for the same task.

A real example of the generated summaries with and without topic control is shown in Table~\ref{tab:generated_summaries} for a super-article that contains a mixture of the {\em transportation} and {\em neuroscience} topics. We notice that the summary of BART discusses only one of the two topics of the super-article, while the control tokens in BART\textsubscript{tag} can successfully shift the generation towards the desired topic of the super-article.

\begin{table}
    \label{tab:generated_summaries}
    \caption{ Summaries generated by BART\textsubscript{tag} according to the two different topics of the super-article along with the generic summary generated by BART.}
    \centering   
    \begin{tabularx}{\columnwidth}{|X|}
    \hline
    \textbf{Generic summary:} Ford unveiled two prototype electric bikes at Mobile World Congress in Barcelona. MoDe:Me and Mo de:Pro are powered by 200-watt motors and fold to fit on a train or in the boot of a car With pedal assist they help riders reach speeds of up to 15mph (25km/h) The bikes are part of an experiment by Ford called Handle on Mobility.  \\ \\
    \textbf{Transportation:}  Ford unveiled two prototype electric bikes at Mobile World Congress in Barcelona. The MoDe: Me and Mo de: Pro are powered by 200-watt motors. They fold to fit on a train or in the boot of a car.With pedal assist, riders reach speeds of up to 15mph (25km/h) \\ \\
    \textbf{Neuroscience:} Researchers from Bristol University measured biosonar bat calls to calculate what members of group perceived as they foraged for food. Pair of Daubenton’s bats foraged low over water for stranded insects at a site near the village of Barrow Gurney, in Somerset. It found the bats interact by swapping between leading and following, and they swap these roles by copying the route a nearby individual was using up to 500 milliseconds earlier.\\ \hline
    \end{tabularx}

\end{table}

\subsection{Zero-shot Experimental Evaluation}
In contrast to the embedding-based models, all the methods that use control tokens can directly handle unknown topics. More specifically, for the prepending method we simply prepend the unknown topic to the document while for the tagging method we tag the most representative words for the unknown topic, assuming the existence of a representative set of documents for this topic. To demonstrate the efficacy of control tokens on unseen topics, we fine-tune the BART model on the same training set of the created topic-oriented dataset but removing 5\% of the topics. More specifically, we randomly remove 3 topics (i.e., ``\textit{Movies}'', ``\textit{Transportation}'' and ``\textit{Podcasts}'') out of the 70 topics of the training set and evaluate the models on the zero-shot test, which consists of 264 articles of unseen topics, as shown in Table~\ref{tab:zero_shot}. We also employ an LLM (GPT-3.5), prompting it to summarize articles on the requested topic given the prompt ``Summarize the following article for the topic [Topic]''.

\begin{table}
\centering
\label{tab:zero_shot} 
\caption{Results on the topic-oriented CNN/DailyMail test set with unseen topics.}
\begin{tabular}{ccccc}
\toprule
     & \textbf{R-1} & \textbf{R-2}   & \textbf{R-L} & \textbf{STAS (\%)}\\  \midrule
GPT-3.5                 & 24.43 & 6.19  & 14.8   &  58.16 \\
BART\textsubscript{tag} & 37.52 & 16.99 &  35.58 & 74.80 \\
BART\textsubscript{pre} & 38.13 & 17.84 & 35.69  & 74.67  \\
BART\textsubscript{pre+tag} & \textbf{39.22} & \textbf{18.81} &  \textbf{36.81} & \textbf{77.94} \\\midrule
\end{tabular}
\end{table}

Even though the models have not seen the zero-shot topics during training, they can successfully generate topic-oriented summaries for these topics achieving similar results in terms of both ROUGE-1 score and STAS metric, with the BART\textsubscript{pre+tag} method outperforming all the other methods. In addition, the results indicate that all the proposed BART models outperform GPT-3.5, with BART\textsubscript{pre+tag} achieving 39.22 compared to 24.43 ROUGE-1. In addition, the proposed method achieves a significantly higher STAS score ($\sim$78\%) compared to 58.16\% of GPT-3.5. This finding further confirms the capability of methods that use control tokens to generalize successfully to unseen topics, paired with increased efficiency (406M parameters for BART-large vs 175B parameters of GPT-3.5).

\begin{table}[ht!]
\centering
\label{tab:experimental_results_real} 
\caption{Results on the original CNN/DailyMail test set, with oracle and non-oracle guidance.}
\begin{tabular}{cccccc}
\toprule  & \multicolumn{4}{c}{\textbf{Oracle}} & \multicolumn{1}{c}{\textbf{Non-oracle}} \\

& \textbf{R-1} & \textbf{R-2}   & \textbf{R-L} &  \textbf{STAS(\%)} & \textbf{STAS(\%)} \\ \midrule
BART\textsubscript{emb} & 42.93 & 20.27 & 40.14 & 71.14 & 66.76 \\ \midrule
BART\textsubscript{tag} & 42.54 & 20.11 & 39.80 & 71.51 & 67.09 \\ \midrule 
BART\textsubscript{prepend}  & 42.75 & 20.20 & 39.94 & 74.20 & 69.67 \\ \midrule 
BART\textsubscript{prepend+tag}  & \textbf{43.35} & \textbf{20.66} & \textbf{40.53} & \textbf{74.23} & \textbf{70.09} \\ \bottomrule
\end{tabular}
\end{table}

\subsection{Experimental Results on Original CNN/DailyMail}
We evaluate on the original CNN/DailyMail test set all the proposed methods fine-tuned on the topic-oriented CNN/DailyMail training and validation sets, using both an oracle setup, where the topic information is extracted from the target summary according to the assigned topical dataset, and a non-oracle setup, where the topic information is extracted directly from the input document. More specifically, for the non-oracle setup, we extract the top-3 topics from the input article. For computational reasons, we sample 3,000 articles from the test set of the original CNN/DailyMail and we predict the summary for each of the three different topics. STAS is therefore computed on these 9,000 pairs of topics and articles. 

The results are shown in Table~\ref{tab:experimental_results_real}. All models perform quite similarly in terms of ROUGE score in the oracle setup, while the best performance is achieved when tagging is combined with prepending, outperforming all the evaluated methods. We do not compute ROUGE scores in the non-oracle setup, as we lack a gold summary for each different topic in this case. 

In terms of STAS, in both setups prepending leads to much better results compared to token embeddings and tagging, which have similar scores. The best results are again obtained when tagging is combined with prepending. The high STAS score of the combined BART\textsubscript{pre+tag} model in the non-oracle (70.09\%) setup shows that this model can successfully shift the generation towards multiple different topics.


\subsection{Human Evaluation}
In order to validate the reliability of STAS, we conducted a human evaluation study. More specifically, we retrieved a set of 83 summary-topic pairs. Then, we asked 80 volunteer participants, including both graduate and undergraduate students, to participate in this evaluation study. Our main objective was to gauge how well STAS is correlated with human judgments. We asked participants to evaluate how relevant is the generated summary with respect to the given topic. The human evaluation was presented as a multiple-choice question with 10 answers (1 to 10). Each time, we picked either a relevant or an irrelevant topic for the given summary. Each summary topic-pair was annotated by an average of 2.8 annotators. Inter-annotations with a more than 5-degree variance were subjected to manual evaluation and discarded. The inter-annotator agreement across the raters for each sample was 0.86 and measured using Krippendorff’s alpha coefficient~\cite{krippendorff2004content}, with ordinal weights~\cite{hayes2007answering}. The high inter-annotator agreement is an indicator of consistency across different raters.

\begin{table}
    \centering
    \caption{Correlation between human evaluation and STAS measure.}
    \label{tab:correlations} 
    \begin{tabular}{ccc}
    \toprule
     \textbf{Metric} & \textbf{Correlation} & \textbf{p-value} \\ \midrule 
    Pearson         & 0.94 &  7.8e-63\\ 
 
    Spearman     & 0.87 & 8.4e-63 \\ 
    \bottomrule
    \end{tabular}
\end{table}

We also investigate the score ranges between human annotators and the STAS metric to better interpret the STAS results. The 10-point evaluation scale allows for better interpretation of the STAS scores, especially in cases where the topic might be relevant but not dominant in the given summary. At the same time, this also gives us an estimation of how to interpret STAS scores on a 1–10 scale based on the human annotations. Thus, we indicate a minimum threshold of the STAS metric for a summary to be strongly related to a topic. More specifically, if a human annotator evaluates the summary for a topic with more than 8, we suppose that this topic is also strongly included in the summary. Thus, for all the human evaluations with a score equal to or higher than 8, we take the minimum value of STAS, i.e., 69.60\%,to indicate the minimum threshold of the STAS metric for actively discussing a topic in the summary.

\section{Conclusions and Future Work}
\label{sec:conclusion}
We proposed STAS, a structured way to evaluate the generated summaries. In addition, we conducted a user study to validate and interpret the STAS score ranges. We also proposed topic-controllable methods that employ either topic embeddings or control tokens demonstrating that the latter can successfully influence the summary generation towards the desired topic, even in a zero-shot setup. Our empirical evaluation further highlighted the effectiveness of control tokens achieving better performance than embedding-based methods, while being significantly faster and easier to apply.

Future research could examine other controllable aspects, such as style~\cite{fan2018controllable}, entities~\cite{chan2021controllable} or length~\cite{liu2020,chan2021controllable}. In addition, the tagging-based method could be further extended to working with any arbitrary topic, bypassing the requirement of having a labeled document collection of a topic to guide the summary towards this topic. Finally, richer vector representations, such as contextual embeddings, could be explored to further improve the performance of the proposed methods.

\bibliographystyle{unsrt}  
\bibliography{references, references1}

\end{document}